\documentclass{article}
\usepackage[utf8]{inputenc}

\title{Synthetic-to-Real Unsupervised Domain Adaptation for Scene Text Detection in the Wild}
\author{Weijia Wu$^1$ , Ning Lu$^2$ , Enze Xie$^3$,  Hong Zhou$^1$}
\date{$^1$ Zhejiang University \\
$^2$ Tencent\\
$^3$ University of Hong Kong.}

\usepackage{graphicx}
\usepackage{graphicx}
\usepackage{amsmath,amssymb} % define this before the line numbering.

\usepackage{color}
\usepackage{multirow}
\usepackage{makecell}
\usepackage[ruled]{algorithm2e}
\usepackage{graphicx} %插入图片的宏包
\usepackage{float} %设置图片浮动位置的宏包
\usepackage{subfigure} %插入多图时用子图显示的宏包

\def\ie{\emph{i.e.,}}

\begin{document}

\maketitle
\begin{abstract}
Deep learning-based scene text detection can achieve preferable performance, powered with sufficient labeled training data. However, manual labeling is time consuming and laborious. At the extreme, the corresponding annotated data are unavailable. Exploiting synthetic data is a very promising solution except for domain distribution mismatches between synthetic datasets and real datasets. To address the severe domain distribution mismatch, we propose a synthetic-to-real domain adaptation method for scene text detection, which transfers knowledge from synthetic data (source domain) to real data (target domain). In this paper, a text self-training (TST) method and adversarial text instance alignment (ATA) for domain adaptive scene text detection are introduced. ATA helps the network learn domain-invariant features by training a domain classifier in an adversarial manner. TST diminishes the adverse effects of false positives~(FPs) and false negatives~(FNs) from inaccurate pseudo-labels. Two components have positive effects on improving the performance of scene text detectors when adapting from synthetic-to-real scenes. We evaluate the proposed method by transferring from SynthText, VISD to ICDAR2015, ICDAR2013. The results demonstrate the effectiveness of the proposed method with up to \textbf{$10\%$} improvement, which has important exploration significance for domain adaptive scene text detection. Code is available at \url{https://github.com/weijiawu/SyntoReal_STD}
\end{abstract}

%===========================================================
\section{Introduction}
Scene text detection has received increasing attention due to its numerous applications in computer vision. Additionally, scene text detection~\cite{he2017deep,liao2017textboxes,liu2019towards,long2018textsnake,tian2016detecting,wang2019shape,zhou2017east} has achieved great success in the last few decades. However, these detection methods require manually labeling large quantities of training data, which is very expensive and time consuming. Whereas several public benchmarks~\cite{karatzas2013icdar,karatzas2015icdar,nayef2017icdar2017,yuan2018chinese,yao2012detecting} have already existed, they only covered a very limited range of scenarios. In the real world, a specific application task usually requires the collection and annotation of a new training dataset, and it is difficult, even impossible, to collect enough labeled data. Therefore, the expensive cost of labeling has become a major problem for text detection applications based on deep learning methods.

\begin{figure*}[t]
	\centering
	\includegraphics[width=1\textwidth]{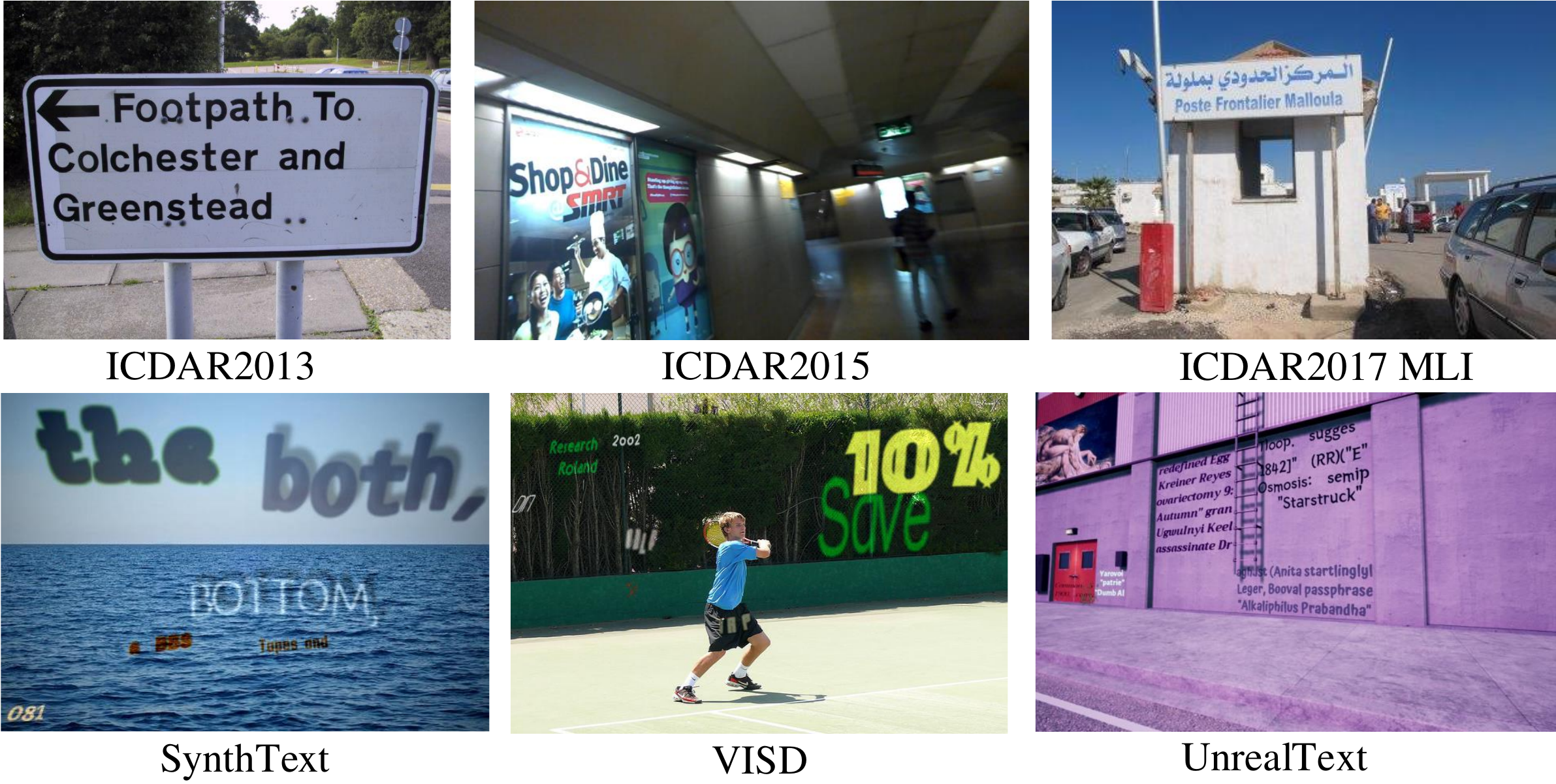}
	\caption{Examples of different datasets. The first row are from real ICDAR2013\cite{karatzas2013icdar}, ICDAR2015\cite{karatzas2015icdar}, and ICDAR2017 MLI\cite{nayef2017icdar2017}, respectively. The second row is from Virtual SynthText\cite{gupta2016synthetic}, VISD\cite{zhan2018verisimilar}, and UnrealText\cite{long2020unrealtext}. There remains a considerable domain gap between synthetic data and real data.}
	\label{fig1}
\end{figure*}

With the great development of computer graphics, an alternative way is to utilize synthetic data, which is largely available from the virtual world, and the ground truth can be freely and automatically generated. SynthText~\cite{gupta2016synthetic} first provides a virtual scene text dataset and automatically generates synthetic images with word-level and character-level annotations. Zhan $et al.$~\cite{zhan2018verisimilar} equipped text synthesis with selective semantic segmentation to produce more realistic samples. UnrealText~\cite{long2020unrealtext} provides realistic virtual scene text images via a 3D graphics engine, which provides realistic appearance by rendering scene and text as a whole. Although synthetic data offer the possibility of substituting for real images in training scene text detectors with low annotation cost and high labeling accuracy, many previous works have also shown that training with only synthetic data degrades the performance on real data due to a phenomenon known as "domain shift". As shown in Fig.~\ref{fig1}, unlike common objects, text has more diversity of shapes, colours, fonts, sizes, and orientations in real-world scenarios, which causes a large domain gap between synthetic data and real data. Therefore, the performance of the model degrades significantly when applying model learning only from synthetic data to real data.

To tackle the domain shift, we propose a synthetic-to-real
domain adaptation approach for scene text detection, which aims to efficiently improve the model performance on real data by using synthetic data and unlabeled real data. Inspired by~\cite{kim2019self} and~\cite{ganin2014unsupervised}, a text self-training(TST) method and an adversarial text instance alignment(ATA) are proposed in our paper to reduce the domain shift. Self-training has achieved excellent results for domain adaptive common object detection~\cite{rosenberg2005semi,roychowdhury2019automatic} and semantic 
segmentation~\cite{sankaranarayanan2018learning,zou2018unsupervised}. However, scene text detection tasks with more diverse situations and complex backgrounds have not been explored in this direction to the best of our knowledge. To better apply self-training to scene text detection, TST is used to suppress the adverse impact of both false positives and false negatives that occur in pseudo-labels. In addition, we first utilize adversarial learning help the model to learn discriminative features of scene text. Adversarial learning has been shown to be effective in tasks such as domain adaptive image classification~\cite{saito2018open} and common object detection~\cite{chen2018domain,zhu2019adapting}. Because most scene text detectors are one-stage detectors since they do not have region proposal process, we propose ATA to align discriminative features for text instances in an adversarial training manner. The contributions of our paper are as follows:

\begin{itemize}
	\item[$\bullet$] We introduce text self-training (TST) to improve the performance of domain adaptive scene text detection by minimizing the adverse effects of inaccurate pseudo-labels.
	\item[$\bullet$] We propose adversarial text instance alignment (ATA) to help the model learn domain-invariant features, which enhance the generalization ability of the model.
	\item[$\bullet$] We first introduce a synthetic-to-real domain adaptation method for scene text detection, which transfers knowledge from the synthetic data (source domain) to real data (target domain).
\end{itemize}

The proposed method is evaluated by extensive experiments for the scene text detection transfer task({\it e.g., SynthText~\cite{gupta2016synthetic}$\rightarrow$ICDAR2015~\cite{karatzas2015icdar}}). The experimental results demonstrate the effectiveness of the proposed approach for addressing the domain shift of scene text detection, which has important exploration significance for domain adaptive scene text detection.

%------------------------------------------------------------------------- 

\section{Related Work}

\subsection{Scene Text Detection}
Before the era of deep learning, SWT~\cite{epshtein2010detecting} and MSER ~\cite{neumann2010method} were two representative algorithms for conventional text detection methods. SWT offers an edge map to obtain information about the text stroke efficiently, and MSER draws intensity stable regions as text candidates.
Based on object detection and semantic segmentation knowledge, scene text detection~\cite{liao2018rotation,xu2019textfield,tian2019learning,tian2016detecting,shi2017detecting} has made great progress. Textboxes++~\cite{liao2018textboxes} adjusts convolutional kernels and anchor boxes to capture various text shapes. EAST~\cite{zhou2017east} performs very dense predictions that are processed using locality-aware NMS. Other methods also draw inspiration from semantic segmentation and detect texts by estimating word bounding areas. PixelLink~\cite{deng2018pixellink} detects text instances by linking neighbouring pixels. PSENet~\cite{li2018shape} proposed a progressive scale algorithm to gradually expand the predefined kernels for scene text detection. The above methods require large-scale manually labeled data.
In addition to the above methods based on strongly supervised learning, some weakly/semi-supervised methods are proposed to reduce the expansive cost of annotation. WeText~\cite{Alpher27} trains a text detection model on a small amount of character-level annotated text images, followed by boosting the performance with a much larger amount of weakly annotated images at word/text line level. WordSup~\cite{WordSup} trains a character detector by exploiting word annotations in rich, large-scale real scene text datasets. ~\cite{wu2020texts} utilizes the network pretrained on synthetic data with full masks to enhance the coarse masks in a real image.

\subsection{Domain Adaptation}
Domain adaptation reduces the domain gap between training and testing data. Prior works~\cite{gong2012geodesic} estimated the domain gap and minimized it. Recent methods use more effective methods to reduce the domain gap, such as incorporating a domain classifier with gradient reversal~\cite{ganin2014unsupervised}. ~\cite{chen2018domain} addressed the domain shift by training domain discriminators on the image level and instance level. ~\cite{kim2019self} introduced a weak self-training to diminish the adverse effects of inaccurate pseudo-labels, and designed adversarial background score regularization to extract discriminative features. For scene text, the domain adaptation method~\cite{zhan2019ga} converts a source-domain image into multiple images of different spatial views as in the target domain. Handwriting recognition~\cite{bhunia2019handwriting} proposes AFDM to elastically warp extracted features in a scalable manner.

\subsection{Self-Training}
Prior works used self-training~\cite{lee2013pseudo,choi2019pseudo} to compensate
for the lack of categorical information. ~\cite{chen2011co} bridged the gap between the source and target domains by adding both the target features and instances in which the current algorithm is the most confident. ~\cite{saito2017asymmetric} used three networks asymmetrically, where two networks were used to label unlabeled target samples and one network was trained to obtain discriminative representations. Other works~\cite{shu2018dirt,zhang2018collaborative,zou2018unsupervised} also showed the effectiveness of self-training for domain adaptation. However, text detection still requires further exploration in the self-training method due to a lack of previous work.
%--------------------------------------------------------------------
\section{Proposed Method}
In this section, the problems caused by domain shifts are analysed. Furthermore, we introduce the principle of TST and ATA, and how to use them for domain adaptation. To evaluate our method, EAST~\cite{zhou2017east} is adopted as the baseline.
\subsection{Problem and Analysis}
 Although synthetic scene text data can be automatically generated with diversified appearance and accurate ground truth annotations, the model trained with only synthetic data cannot be directly applied to real scenes since there exists a significant domain shift between synthetic datasets and real datasets.  
 
 Viewing the problem from a probabilistic perspective is clearer. We refer to the synthetic data domain as the source domain and the real data domain as the target domain. The scene text detection problem can be viewed as learning the posterior $P(B|I)$, where $I$ refers to the image features and $B$ is the predicted bounding-box of text instances. Using the Bayes formula,
the posterior $P(B|I)$ can be decomposed as:
 \begin{equation}
  P(B|I)= \frac{P(I|B)*P(B)}{P(I)} =  \frac{P(I|B)}{P(I)}*P(B)\,.
\end{equation}
 We make the covariate shift assumption in this task that the priori probability
 $P(B)$ is the same for the two domains. $P(I|B)$ refers to the conditional probability of $I$, which means that the likelihood of learning true features given that the predicted result is true. We also consider that $P(I|B)$ is the same for both domains.
 Therefore, the difference in posterior probability is caused by the priori probability $P(I)$. In other words, to detect text instances, the difference in detection results is caused by domain change features. To improve the generalization ability, the model should learn more domain-invariant features, keeping the same $P(I)$ regardless of which domain the input image belongs. 
 
 In the EAST~\cite{zhou2017east} model, the image feature $P(I)$ refers to the features output from the backbone. Therefore, the feature map should be aligned between the source domain and the target domain~(\ie{~$P_s(I)=P_t(I)$}). To achieve this goal, ATA is proposed to align the features, with more details in the next subsection.
 
\begin{figure*}[t]
	\centering
	\includegraphics[width=1\textwidth]{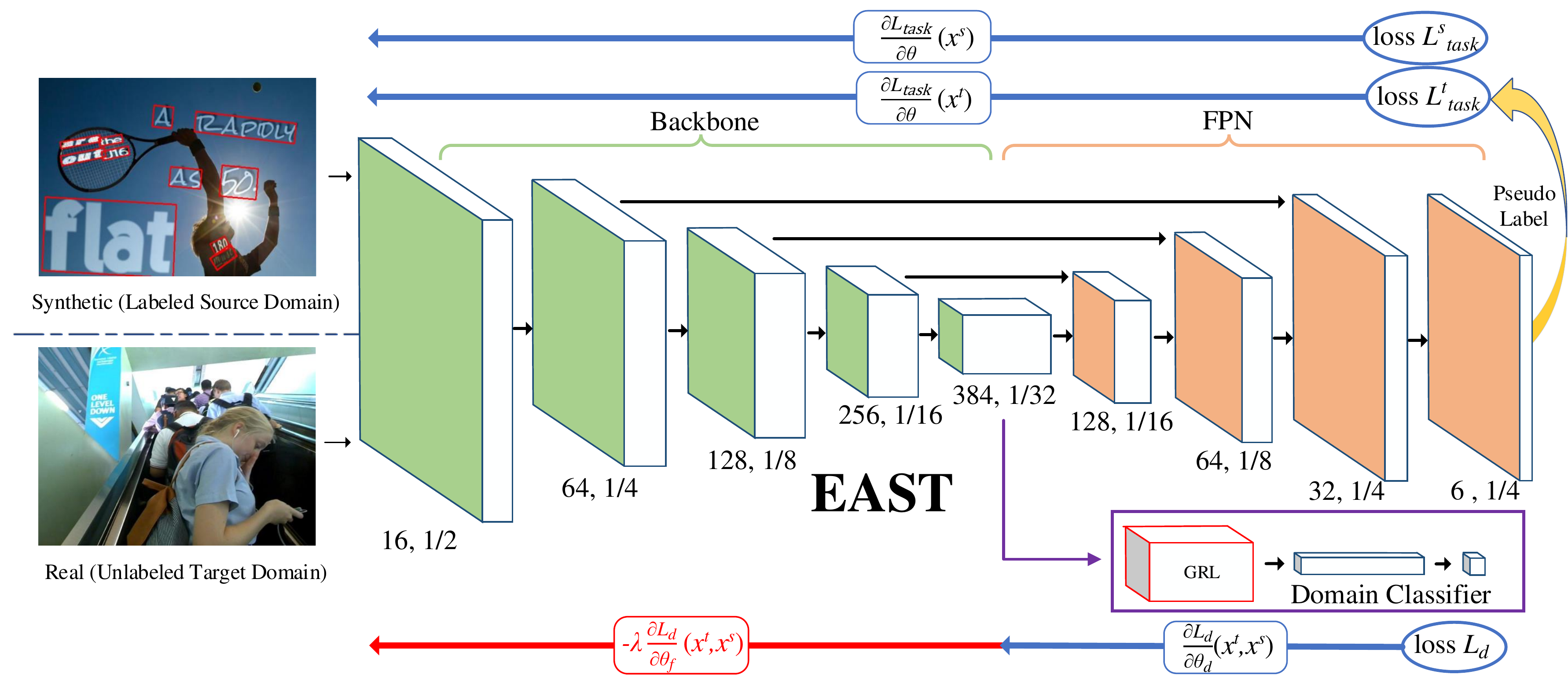}
	\caption{The network architecture with the corresponding optimization object. $\theta$ represents the parameters of EAST. A domain classifier (green) is added after the feature extractor via a gradient reversal layer that multiplies the gradient by a certain negative constant during the backpropagation-based training. $L_{task}$ refers to the original detection loss of EAST, and $L_d$ is the loss of domain classifier.}
	\label{fig2}
\end{figure*}

\subsection{Adversarial Text Instance Alignment }
Motivated by~\cite{ganin2014unsupervised}, ATA is adopted to help the network learn domain-invariant features. In the EAST model, the image features $P(I)$ refer to the feature map outputs of the backbone~(\ie{~384, 1/32 features in Fig.~\ref{fig2}}). To align the features $P(I)$ between the source domain and target domain, a domain classifier is used to confuse the feature domain.

In particular, the domain classifier is trained for each input image and predicts the domain label to which the image belongs. We assume that the model works with input samples $x  \in  X$, where $X$ is the some input space. $y_i$ denotes the domain label of the $i$-th training image, with $y_i$ = 0 for the source domain and $y_i$ = 1 for the target domain. $p_i(x)$ is the output of the domain classifier, and we use cross entropy as the loss function:

\begin{equation}
L_{d} = -\sum_{i}^{}(y_i\times ln^{p_i(x)} + (1-y_i)\times ln^{1 - p_i(x)})\,.
\end{equation}

To learn domain-invariant features, we optimize the parameters in an adversarial way. The parameters of the domain classifier are optimized by minimizing the above domain classification loss, and the parameters of the base network are optimized by maximizing this loss. For more detail, the gradient reverse layer (GRL)~\cite{ganin2014unsupervised} is added between the backbone of EAST and the domain classifier, and the sign of the gradient is reversed when passing through the GRL layer.

As shown in Fig.~\ref{fig2}, both the feature pyramid network(FPN) and the backbone minimize the original loss $L_{task}$ of EAST at the training phase. $L_{task}$ specifically denotes the score map loss and geometries loss in EAST~\cite{zhou2017east}. $L^t_{task}$ refers to training with the pseudo-label in the target domain, and $L^s_{task}$ denotes training with the source domain. Thus, different training objectives for various parameter spaces:

\begin{equation}
\begin{cases}
L_f = min(L^t_{task}(\theta_f|x^t) + L^s_{task}(\theta_f|x^s) - \lambda  L_d(\theta|(x^s,x^t))) \quad & \theta_f\in F\,,  \\ 
L_d = min(L_d(\theta_d|(x^s,x^t))) \quad & \theta_d\in C\,,  \\ 
L_h = min(L^t_{task}(\theta_h|x^t) + L^s_{task}(\theta_h|x^s)) \quad & \theta_h\in D \,,
\end{cases}
\end{equation}
where $F,C,D$ are the parameter spaces of the backbone, the domain classifier and the FPN. The overall training objective is as follows:
\begin{equation}
L= L_f + L_h + \lambda L_d \,,
\end{equation}
where $\lambda$ is the tradeoff parameter, we set it to 0.2 in all experiments. Through optimizing the loss, the network can learn more text domain-invariant features, transforming better from synthetic data to real data.

\subsection{Text Self-Training}
Previous works~\cite{zou2018unsupervised,inoue2018cross} have shown the effectiveness of self-training for domain adaptation. However, two major problems for self-training still need to be explored further: false positives(FP) and false negatives(FN) occurred in pseudo-label. Incorrect pseudo-labels will cause very serious negative effects to our networks. To overcome such problems, TST is designed to minimize the adverse effects of FP and FN.

\subsubsection{Reducing False Negatives.} Inspired by~\cite{kim2019self}, a weak supervision way is utilized to minimize the effects of false negatives. As defined in EAST~\cite{zhou2017east}, the original score map loss is

\begin{equation}
\begin{aligned}
L_s  =-\sum_{i\in Pos}^{}\beta Y^*log \widehat{Y}-\sum_{i\in Neg}^{}(1-\beta)(1-Y^*)(1-\widehat{Y})\,,
\end{aligned}
\end{equation}
where $\widehat{Y}=F_s$ is the prediction of the score map, and $Y^*$ is the ground truth. While the network is optimized by backpropagation learning the loss of background~(\ie{~negative examples}), FP occurring in pseudo-labels misleads the network. We assume that FPs are mainly selected by hard negative mining, such as blurred text and unusual fonts similar to the background. To reduce the adverse effects of FP, we ignore some background examples that have the potential to be foregrounds with a confidence score.
\begin{figure*}[t]
	\centering
	\includegraphics[width=1\textwidth]{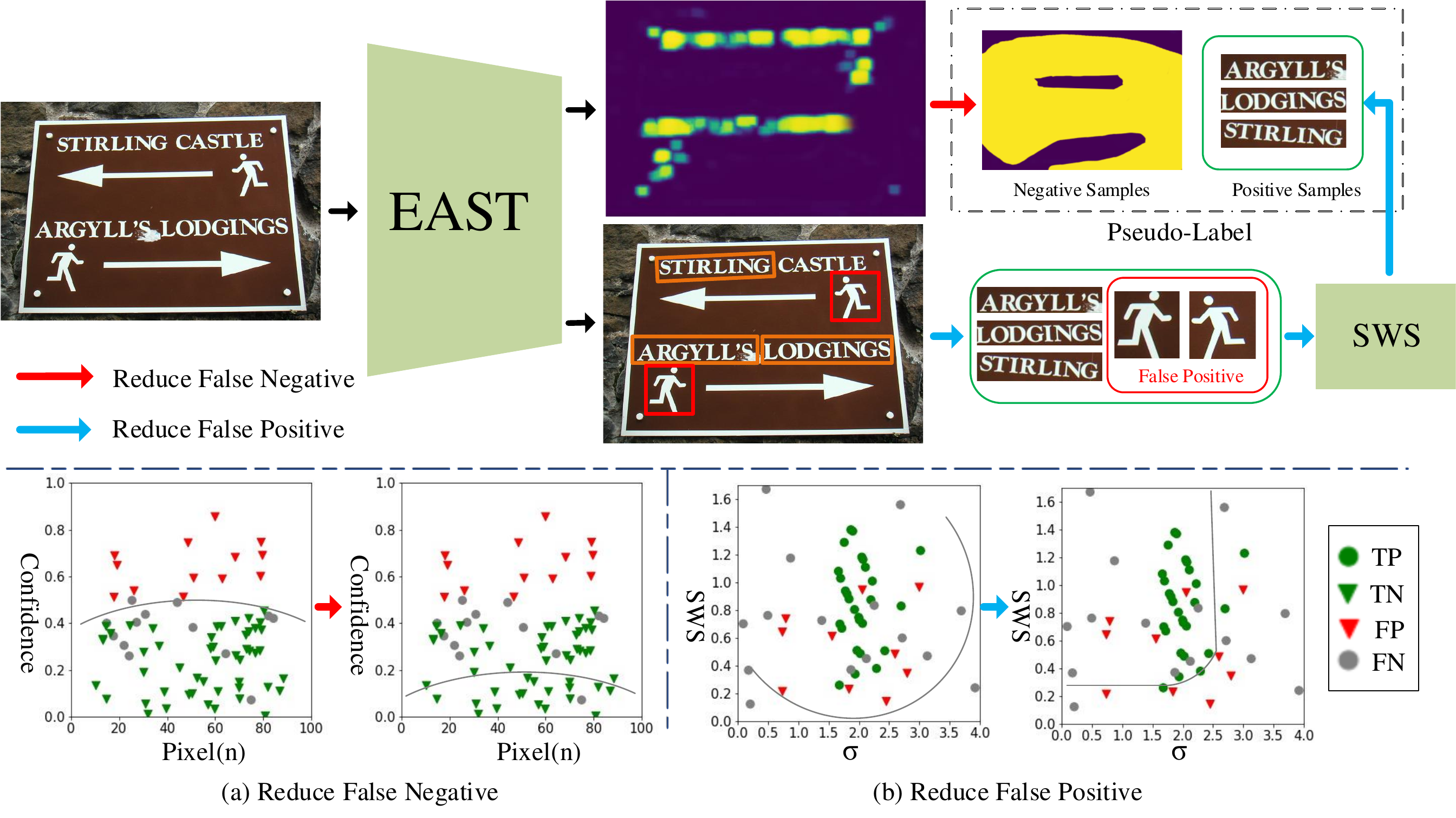}
	\caption{Up: The framework of proposed text self-training. We utilize SWS to filter the positive samples in pseudo-label for minimizing false positives, and select a third of negative samples with low confidence as the final negative samples to minimize false negatives. Down: We present sample space representation for pseudo-label. (a): False negatives are effectively filtered out by weak training. (b): False positives are filtered out by the standard deviation($\sigma$) of the stroke width and SWS. }
	\label{fig3}
\end{figure*}

Negative examples for EAST are a pixel map, a pixel is more likely to be considered a negative pixel while the corresponding confidence score higher. Thus, we choose a part of the negative sample pixels({\it e.g., $Neg/3$}) that have the lowest confidence score as the final negative examples, which is denoted as $\widehat{Neg}$ in Fig.~\ref{fig3}(red line). The corresponding mathematical expression is $\widehat{Neg} = \eta Neg$, where $\eta$ is set to $1/3$ in all experiments. For those pixels that have a high confidence score, the network does not optimize this part loss. Finally, the modified loss function is defined as

\begin{equation}
\begin{aligned}
L_{sw} =-\sum_{i\in Pos}^{}\beta Y_i^*log \widehat{Y}-\sum_{i\in \widehat{Neg}}^{}(1-\beta)(1-Y^*_i)(1-\widehat{Y}) \,.
\end{aligned}
\end{equation}

\subsubsection{Reducing False Positives.}

Corresponding to false negatives, false positives also cause serious interference to the network. Some patterns and designs in natural scenes are extremely easy to identify as text, which leads to inaccurate pseudo-labels. Replacing Supporting Region-based Reliable Score(SRRS) in~\cite{kim2019self}, we propose a more reasonable Stroke Width Score(SWS) that utilizes the Stroke Width Transform(SWT)~\cite{epshtein2010detecting} to evaluate the predicted boxes of text instances. On the one hand, SRRS is not applicable to EAST based on segmentation. SRRS in~\cite{kim2019self} is define as:
% \begin{small}
\begin{equation}
  SRRS(r^*)= \frac{1}{N_s}\sum_{i=1}^{N_s}IoU(r_i,r^*)\cdot P(c^*|r_i)
\end{equation}
% \end{small}
The EAST is a segmentation-based method without FPN, the text instances with small area have less supporting boxes~(\ie{$r_i$}) than that with big area, which leads to the extremely unbalanced supporting boxes number~(\ie{$N_s$}). On the other hand, SWT is more reasonable for eliminating non-text regions, and similar previous works~\cite{ozgen2018text,yao2012detecting} have shown its effectiveness.

SWT is a local image operator that computes the most likely stroke width for each pixel. The output of the SWT is a $n*m$ matrix where each element contains the width of the stroke associated with the pixel. Specifically, each pixel on the boundary of the stroke is connected with the opposite side of the stroke in the direction of the gradient, and the width of the connecting line is the width of the stroke for the pixel. SWS assesses the predicted boxes by utilizing the information of the corresponding stroke width, and eliminates part of the non-text regions, as shown in Fig.~\ref{fig3}(blue line).

For a typical text region, the variance in stroke width is low since text tends to maintain a fixed stroke width. We denote the set of stroke widths in the $v_{th}$ predicted box as $W_n^v$ and the stroke width of the $u_{th}$ pixel as $w_u^v \in W_n^v$. The standard deviation is as follows:

\begin{equation}
\sigma_v = \sqrt{\frac{1}{N}\sum_{u=1}^{N}(w_u^v-\mu^v )^2}\,,
\end{equation}
where $\mu^v$ is the mean stroke width in the $v_{th}$ predicted box
Therefore, each predicted box has a standard deviation($\sigma$) about the stroke width, and we choose reliable boxes with an upper threshold($\epsilon_1 $). In addition, we further filter the boxes by SWS:
\begin{equation}
SWS_{v} = \frac{w_v}{\sigma_v^2}\,,
\end{equation}
$w_v$ is the most common stroke width value for the $v_{th}$ predicted box. By thresholding the score with a lower threshold $\epsilon_2 $, the boxes are further selected. Fig. 3 (b) shows that part of the FP is filtered out by SWS and $\sigma$.

%--------------------------------------------------------------------
\section{Experiments}
The proposed method is evaluated by transferring a scene text detector from synthetic datasets to real datasets.  We adopt several pure synthetic data and real scene data~(\ie{~SynthText~\cite{gupta2016synthetic} and ICDAR2015~\cite{karatzas2015icdar}}), which have common annotation style~(\ie{~word level}).

\subsection{Dataset and Experimental Settings}

\subsubsection{Pure Synthetic Datasets.} \emph{SynthText}~\cite{gupta2016synthetic} is a large-scale dataset that contains about 800K synthetic images. These images are created by blending natural images with text rendered with random fonts, sizes, colours, and orientations, thus these images are quite realistic. ~\emph{Verisimilar Image Synthesis Dataset(VISD)}~\cite{zhan2018verisimilar} contains 10~k images synthesized with 10~k background images. Thus, there are no repeated background images for this dataset. The rich background images make this dataset more diverse.

\begin{table}[t]
\fontsize{8}{10}\selectfont
\begin{center}
\caption{
The performance of different models on Syn2Real scene text detection dataset for SynthText/VISD$\rightarrow$ICDAR2015 transfers. UL refers to the unlabeled data. * denotes the performance reported in UnrealText~\cite{long2020unrealtext}. $\dagger$ refers to our testing performance.
}
\label{table1}
\centering
\begin{tabular}{|c|c|c|c|c|c|}
\hline
\multirow{2}*{\makecell[c]{Method}}  & \multirow{2}*{\makecell[c]{Source $\rightarrow$ Target(UL)}} & \multirow{2}*{Annotation} &  \multicolumn{3}{c|}{Detection Evaluation/\%}\cr\cline{4-6}

~ &  ~ & ~ & Precision & Recall & F-score \cr\hline
PAN~\cite{wang2019efficient}& SynthText$\rightarrow$ ICDAR2015 & Word & 0.659 & 0.469&0.548 \cr\hline
$EAST^*$~\cite{zhou2017east}& SynthText$\rightarrow$ ICDAR2015 & Word & - & - &0.580 \cr\hline
$EAST^{\dagger}$~\cite{zhou2017east} & SynthText $\rightarrow$ ICDAR2015 & Word & 0.721 & 0.521 &0.605 \cr\hline
CCN~\cite{xing2019convolutional}& SynthText $\rightarrow$ ICDAR2015 & Character & - & -&0.651 \cr\hline
EAST+Ours& SynthText $\rightarrow$ ICDAR2015 & Word &0.690&0.670&0.680 \cr\hline\hline
$EAST^*$~\cite{zhou2017east} & VISD $\rightarrow$ ICDAR2015 & Word & - & - &0.643 \cr\hline
$EAST^{\dagger}$~\cite{zhou2017east} & VISD $\rightarrow$ ICDAR2015 & Word & 0.640 & 0.652 &0.645 \cr\hline
EAST+Ours& VISD $\rightarrow$ ICDAR2015 & Word &\textbf{0.748}&\textbf{0.727}&\textbf{0.738} \cr\hline

\end{tabular}
\end{center}
\end{table}

\subsubsection{Real Datasets.}~\emph{ICDAR2015}~\cite{karatzas2015icdar} is a multi-oriented text detection dataset for English text that includes only 1,000 training images and 500 testing images. Scene text images in this dataset were taken by Google Glasses without taking care of positioning, image quality, and viewpoint. This dataset features small, blur, and multi-oriented text instances. \emph{ICDAR2013}~\cite{karatzas2013icdar} was released during the ICDAR 2013 Robust Reading Competition for focused scene text detection, consisting of high-resolution images, 229 for training and 233 for testing, containing texts in English. 

\subsubsection{Implementation Details.}
In all experiments, we used EAST~\cite{zhou2017east} as a base network.
Following the original paper, inputs were resized to $512\times 512$, and we applied all augmentations used in the original paper. The network was trained with a batch input composed of 12 images, 6 images from the source domain, and the other 6 images from the target domain. The Adam optimizer was adopted as our learning rate scheme. All of the experiments used the same training strategy: (1) pretraining the network for 80~k iterations with ATA to learn domain-invariant features and (2) the pretrained model is used to generate corresponding pseudo-label(\ie{pseudo-bounding box label and negative sample map}) for each image in the target domain, then fine-tuning the pretrained model with generated pseudo-labels. In the process of generating pseudo-labels, we set $\epsilon_1 $ and $\epsilon_2 $ to 3.0 and 0.30 for stroke width elimination parameters. All of the experiments were conducted on a regular workstation (CPU: Intel(R) Core(TM) i7-7800X CPU @ 3.50 GHz; GPU: GTX 2080Ti).

\begin{table}[t]
\fontsize{8}{10}\selectfont
\begin{center}
\caption{
The performance of different models on Syn2Real scene text detection dataset for SynthText/VISD$\rightarrow$ICDAR2013 transfers. UL refers to the unlabeled data. * denotes the performance reported in UnrealText~\cite{long2020unrealtext}. $\dagger$ refers to our testing performance.
}
\label{table2}
\centering
\begin{tabular}{|c|c|c|c|c|c|}
\hline
\multirow{2}*{\makecell[c]{Method}}  & \multirow{2}*{\makecell[c]{Source $\rightarrow$ Target(UL)}} & \multirow{2}*{Annotation} &  \multicolumn{3}{c|}{Detection Evaluation/\%}\cr\cline{4-6}

~ &  ~ & ~ & Precision & Recall & F-score \cr\hline
$EAST^*$~\cite{zhou2017east}& SynthText$\rightarrow$ ICDAR2013 & Word & - & - &0.677 \cr\hline
$EAST^{\dagger}$~\cite{zhou2017east} & SynthText$\rightarrow$ ICDAR2013 & Word & 0.669 & 0.674 &0.671 \cr\hline
EAST+Ours& SynthText$\rightarrow$ ICDAR2013 & Word &0.805&0.765&0.784 \cr\hline\hline
$EAST^*$~\cite{zhou2017east} & VISD$\rightarrow$ ICDAR2013 & Word & - & - &0.748 \cr\hline
$EAST^{\dagger}$~\cite{zhou2017east} & VISD$\rightarrow$ ICDAR2013 & Word & 0.783 & 0.705 &0.742 \cr\hline
EAST+Ours& VISD$\rightarrow$ ICDAR2013 & Word &\textbf{0.830}&\textbf{0.781}&\textbf{0.805} \cr\hline
\end{tabular}
\end{center}

\end{table}

\subsection{Performance Comparison and Analysis}

\subsubsection{Synthetic$\rightarrow$ ICDAR2015 Transfer.} Table~\ref{table1} summarizes the performance comparisons for synthetic$\rightarrow$ICDAR2015 transfer task. The EAST model as the baseline training with source-only had an unsatisfactory F-score($60.5\%$ using SynthText and $64.5\%$ using VISD), which can be regarded as a lower bound without adaptation. By combining with the proposed method, the F-score achieved a $68.0\%$ and $73.8\%$ respectively, making $7.5\%$ and $9.3\%$ absolute improvements over the baseline. GCN~\cite{xing2019convolutional} based on character annotation led to a performance improvement over that based on word annotation. However, the performances of GCN were still lower than our method, which utilizes self-training and adversarial learning. The experiment indicates the efficient performance of the proposed method in alleviating the domain discrepancy over the source and target data.

\subsubsection{Synthetic$\rightarrow$ ICDAR2013 Transfer.}
To further verify the effectiveness of our proposed method, we conducted experiments by using ICDAR2013 as the target domain for the synthetic$\rightarrow$real scene text detection transfer task. The experimental results are reported in Table \ref{table2}. Specifically, for the SynthText$\rightarrow$ICDAR2013 transfer task, compared with the baseline EAST training with source-only, we achieved an $11.3\%$ performance improvement. Similar to synthetic$\rightarrow$ICDAR2015 transfer experiment, VISD was also used as the source domain in the comparison experiment. After using ATA and TST, the proposed method achieved a $6.3\%$ performance improvement over the baseline EAST, which exhibits the effectiveness of the method for reducing the domain shift. Note that for fair comparison, except for adding ATA and TST, the base network and experimental settings of the proposed method were the same as the baseline EAST.

\subsubsection{ICDAR2013$\rightarrow$ ICDAR2015 Transfer.}
Table \ref{table3} shows the performance for  ICDAR2013$\rightarrow$ ICDAR2015 Transfer task. The annotations of ICDAR2013 are rectangular boxes while that of ICDAR2015 are rotated boxes, which limits the transfer performance. However, comparing with the baseline EAST training with source-only, we achieved an $7.6\%$ performance improvement. 

\subsection{Ablation Study}
\subsubsection{Component Analysis.}
To verify the effectiveness of the proposed method, we conducted ablation experiments for Syn2Real transfer task on four datasets: SynthText, VISD, ICDAR2015, and ICDAR2013. Table \ref{table3} shows the experimental results. For the SynthText$\rightarrow$ICDAR2015 transfer task, the F-scores increased by $4.1\%$ and $3.5\%$ respectively, after using the TST and ATA. In addition, our method produced a higher recall rate of up to eight percent than the baseline, which shows the effectiveness of this approach on improving the robustness of the model. By combining both components, the F-score of the proposed method achieved a $68.0\%$, a $7.5\%$ absolute improvement over the baseline. The VISD$\rightarrow$ICDAR2015 transfer task exhibited better performance since VISD has a more realistic synthesis effect. In particular, the F-score using our method reached $73.8\%$, making the absolute improvement over the corresponding baseline $9.3\%$. For SynthText/VISD$\rightarrow$ICDAR2015 transfers, the improved performances are also significant. We achieved a $11.3\%$ performance improvement using SynthText and $6.3\%$ performance improvement using VISD.

\subsubsection{Parameter Sensitivity on TST.}
To explore the influence of threshold parameters~(\ie{~$\epsilon_1$ and $\epsilon_2 $}) on SWS, we conducted several sets of comparative experiments, and the results shown are in Table~\ref{table4}. Threshold parameter $\epsilon_1$ was utilized to filter the predicted box since we considered the standard deviation of the stroke width in the text region close to zero in an ideal situation. The network trained with $\epsilon_1= 3 $ showed better performance than the others, and the results were not sensitive to the parameters. Similar to $\epsilon_1$, three different values $0.2,0.3,0.4$ were adopted to verify the parameter sensitivity of $\epsilon_2$, and the result shows that the value($0.3$) of $\epsilon_2$ was reasonable.

\setlength{\tabcolsep}{4pt}

\begin{table}
\begin{center}
\caption{
Ablation study for the proposed Syn2Real scene text detection transfer. 'Baseline' denotes training only with labeled data in the source domain. $\blacktriangle$ denotes the increase in the F-score compared with the baseline training with source-only. UL refers to the unlabeled data. 'F-target' denotes pretrain in source domain and fine-tuning with original pseudo-bounding box in target domain.
}
\fontsize{8}{10}\selectfont
\label{table3}
\centering
\begin{tabular}{|c|c|c|c|c|c|c|c|}
\hline
\multirow{2}*{Method} &\multirow{2}*{TST} & \multirow{2}*{ATA} & \multirow{2}*{\makecell[c]{Source $\rightarrow$ Target(UL)}} &  \multicolumn{3}{c|}{Detection Evaluation/\%}& \multirow{2}*{Improv.}\cr\cline{5-7}
~ & ~  & ~ & ~ & Precision & Recall & F-score& ~ \cr\hline
Baseline& &  & \multirow{5}*{SynthText$\rightarrow$ ICDAR2015} & 0.721 & 0.521&0.605&-\cr
\cline{1-1}
\cline{5-8}
F-target&  & &~ &0.666&0.535&0.594&-\cr

\cline{1-1}
\cline{5-8}
\multirow{3}*{Ours}& \checkmark & &~ &0.693&0.605&0.646&$\blacktriangle$ 4.1\%\cr
\cline{5-8}
&~& \checkmark &~ &  0.682&0.610&0.640&$\blacktriangle$ 3.5\%\cr
\cline{5-8}
& \checkmark & \checkmark ~&~ &0.690&0.670&0.680 &$\blacktriangle$ 7.5\%  \cr\cline{1-4}
\cline{5-8}

Baseline&  & ~& \multirow{4}* { VISD$\rightarrow$ ICDAR2015}  & 0.640 & 0.652&0.645&-\cr
\cline{1-1}
\cline{5-8}
\multirow{3}*{Ours}& \checkmark & &~ &0.702&0.688&0.695&$\blacktriangle$ 5.0\%\cr
\cline{5-8}
  &~ & \checkmark &~&0.713&0.670&0.691&$\blacktriangle$ 4.6\%\cr
\cline{5-8}
& \checkmark & \checkmark ~&~ &\textbf{0.748}&\textbf{0.727}&\textbf{0.738} &$\blacktriangle$ 9.3\% \cr\hline\hline

Baseline&  & ~& \multirow{4}*{SynthText$\rightarrow$ ICDAR2013}  & 0.669 & 0.674&0.671&-\cr
\cline{1-1}
\cline{5-8}
\multirow{3}*{Ours}& \checkmark & &~ &0.715&0.707&0.711&$\blacktriangle$ 4.0\%\cr
\cline{5-8}
  &~ & \checkmark &~&0.736&0.721&0.729&$\blacktriangle$ 5.8\%\cr
\cline{5-8}
& \checkmark & \checkmark ~&~ &0.805&0.765&0.784 &$\blacktriangle$ 11.3\%  \cr\hline

Baseline&  & ~& \multirow{4}*{VISD$\rightarrow$ ICDAR2013}  & 0.783 & 0.705&0.742&-\cr
\cline{1-1}
\cline{5-8}
\multirow{3}*{Ours}& \checkmark & &~ &0.794&0.720&0.755&$\blacktriangle$ 1.3\%\cr
\cline{5-8}
  &~ & \checkmark &~&0.802&0.751&0.776&$\blacktriangle$ 3.4\%\cr
\cline{5-8}
& \checkmark & \checkmark ~&~ &\textbf{0.830}&\textbf{0.781}&\textbf{0.805} &$\blacktriangle$ 6.3\%  \cr\hline

Baseline&  & ~& \multirow{4}*{ICDAR13$\rightarrow$ ICDAR2015}  & 0.513 & 0.398 &0.448&-\cr
\cline{1-1}
\cline{5-8}
\multirow{3}*{Ours}& \checkmark & &~ &0.546&0.459&0.505& $\blacktriangle$    5.7\%\cr
\cline{5-8}
  &~ & \checkmark &~&0.560&0.441&0.493&$\blacktriangle$  4.5\%\cr
\cline{5-8}
& \checkmark & \checkmark ~&~ &\textbf{0.563}&\textbf{0.490}&\textbf{0.524} & $\blacktriangle$ 7.6\%  \cr\hline

\end{tabular}
\end{center}

\end{table}

\subsubsection{Qualitative Analysis.}
\begin{figure*}[t]
	\centering
	\includegraphics[width=1\textwidth]{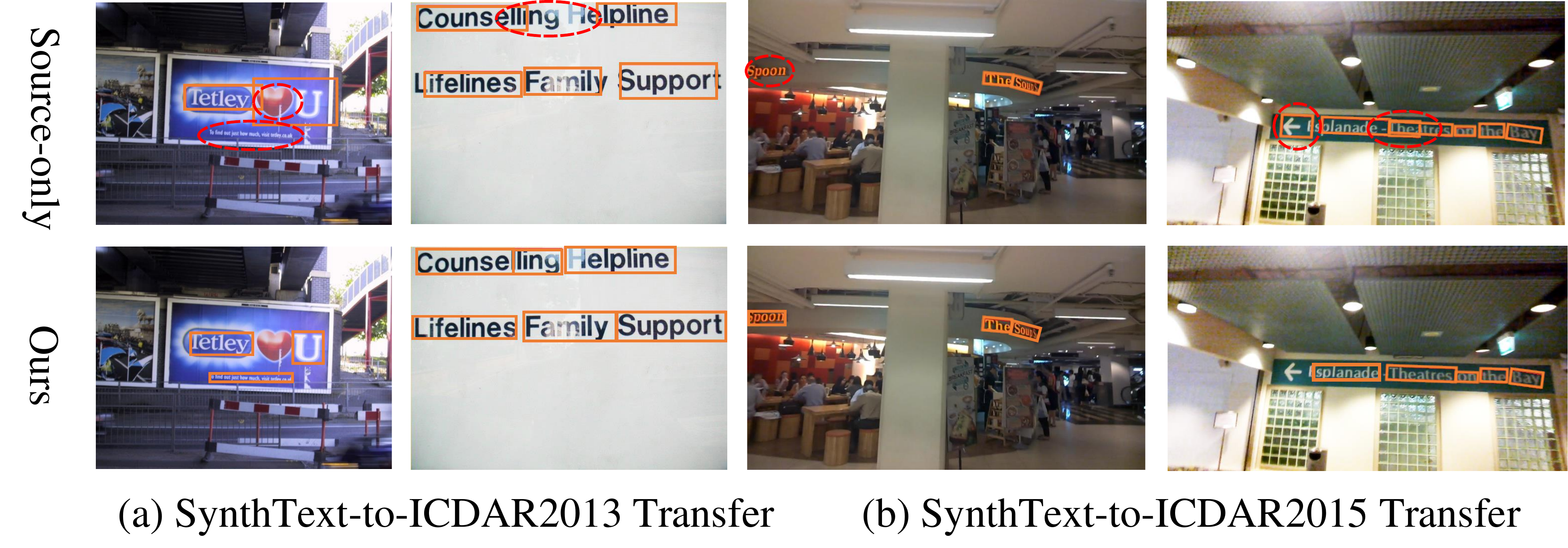}
	\caption{Examples of detection results for different models. The first row is the results of the baseline training with only source domain. The second row is the results of using the proposed method.}
	\label{examples}
\end{figure*}

Fig. \ref{examples} shows four examples of text detection results for synthetic-to-real transfer tasks. The exemplar results clearly show that the proposed method improves the robustness of the model while facing different complex backgrounds and various texts. TST minimizes a part of FP and FN, as shown in the first column example, and ATA helps the model learn more discriminative features. 

\setlength{\tabcolsep}{4pt}
\begin{table}
\fontsize{8}{10}\selectfont
\begin{center}
\caption{
Model results for different values of $\epsilon_1$ and $\epsilon_2$ in Text Self-training(TST).}
\label{table4}
\centering
\begin{tabular}{|c|c|c|c|c|c|c|c|c|c|}
\hline
\multirow{2}*{$\epsilon_2$} &\multirow{2}*{$\epsilon_1$}   &  \multicolumn{3}{c|}{Detection Evaluation/\%} &\multirow{2}*{$\epsilon_1$}& \multirow{2}*{$\epsilon_2 $}  &  \multicolumn{3}{c|}{Detection Evaluation/\%}\cr\cline{3-5}\cline{8-10}

~&~ & Precision & Recall & F-score&~& ~ & Precision & Recall & F-score \cr\hline

\multirow{4}*{-} &- & 0.597 & 0.561 & 0.580 &\multirow{4}*{-} & - & 0.597 & 0.561 & 0.580 \cr\cline{2-5}\cline{7-10}
~ &2 & 0.636&0.543&0.581& ~ & 0.20 & 0.621 & 0.556&0.586\cr\cline{2-5}\cline{7-10}
~ &3 & 0.634&0.563&0.596& ~ & 0.30 & 0.623 & 0.554&0.586\cr\cline{2-5}\cline{7-10}
~ &4 & 0.612&0.565&0.588& ~ & 0.40 & 0.645 & 0.550&0.594\cr\hline
\end{tabular}
\end{center}
\end{table}
\setlength{\tabcolsep}{1.4pt}

\section{Conclusions}
In this paper, we first introduced a synthetic-to-real domain adaptation method for scene text detection, which transfers knowledge from synthetic data (source domain) to real data (target domain). Proposed text self-training (TST) effectively minimizes the adverse effects of false negatives and false positives for pseudo-labels, and the adversarial text instance alignment(ATA) helps the network to learn more domain-invariant features in an adversarial way. We evaluated the proposed method with EAST on several common synthetic and real datasets. The experiments showed that our approach makes a great improvement for synthetic-to-real transfer text detection task.

%===========================================================

\bibliographystyle{plain}
\bibliography{main}
\end{document}